\documentclass[graybox]{svmult}

\usepackage{type1cm}        % activate if the above 3 fonts are
\usepackage{makeidx}         % allows index generation
\usepackage{graphicx}        % standard LaTeX graphics tool
\usepackage{multicol}        % used for the two-column index
\usepackage[bottom]{footmisc}% places footnotes at page bottom

\usepackage{newtxtext}       % 
\usepackage{newtxmath}       % selects Times Roman as basic font

\usepackage{xcolor}

\usepackage[nocompress]{cite}
\makeindex             % used for the subject index
\begin{document}

\title*{Memorability: An image-computable measure of information utility}
%\author{Zoya Bylinskii, Anelise Newman, and Aude Oliva}
\author{Zoya Bylinskii, Adobe Research\\
Lore Goetschalckx, Brown University\\
Anelise Newman, Stanford University\\
Aude Oliva, Massachusetts Institute of Technology
}
\authorrunning{Bylinskii, Goetschalckx, Newman, Oliva}
%\institute{Zoya Bylinskii \at Adobe Research, One Broadway, Cambridge MA 02142, \email{bylinski@adobe.com}
%\and Anelise Newman \at Stanford University, 450 Serra Mall, Stanford CA 94305 \email{anelise@stanford.edu}
%\and Aude Oliva \at MIT, 77 Massachusetts Ave, Cambridge MA 02139 \email{oliva@mit.edu}
\maketitle

\abstract*{The pixels in an image, and the objects, scenes, and actions that they compose, determine whether an image will be memorable or forgettable. While memorability varies by image, it is largely independent of an individual observer. Observer independence is what makes memorability an image-computable measure of information, and eligible for automatic prediction. In this chapter, we zoom into memorability with a computational lens, detailing the state-of-the-art algorithms that accurately predict image memorability relative to human behavioral data, using image features at different scales from raw pixels to semantic labels. We discuss the design of algorithms and visualizations for face, object, and scene memorability, as well as algorithms that generalize beyond static scenes to actions and videos. We cover the state-of-the-art deep learning approaches that are the current front runners in the memorability prediction space. Beyond prediction, we show how recent A.I. approaches can be used to create and modify visual memorability. Finally, we preview the computational applications that memorability can power, from filtering visual streams to enhancing augmented reality interfaces.}

\abstract{The pixels in an image, and the objects, scenes, and actions that they compose, determine whether an image will be memorable or forgettable. While memorability varies by image, it is largely independent of an individual observer. Observer independence is what makes memorability an image-computable measure of information, and eligible for automatic prediction. In this chapter, we zoom into memorability with a computational lens, detailing the state-of-the-art algorithms that accurately predict image memorability relative to human behavioral data, using image features at different scales from raw pixels to semantic labels. We discuss the design of algorithms and visualizations for face, object, and scene memorability, as well as algorithms that generalize beyond static scenes to actions and videos. We cover the state-of-the-art deep learning approaches that are the current front runners in the memorability prediction space. Beyond prediction, we show how recent A.I. approaches can be used to create and modify visual memorability. Finally, we preview the computational applications that memorability can power, from filtering visual streams to enhancing augmented reality interfaces.}

% ####################################################

\section{Introduction}\label{intro}

People are remarkably good at remembering images and their details, even when those images do not hold personal significance or contain recognizable places~\cite{brady2008visual,konkle2010conceptual,konkle2010scene,standing1973learning}. 
Foundational memory studies and the follow-up computational work~\cite{isola2011understanding,isola2014what,isola2011makes} proved that images have intrinsic traits that make some of them memorable and others forgettable. 
The earliest computational studies quickly ruled out low-level features like color and contrast and showed that neither aesthetics nor preference explain memorability ~\cite{isola2011understanding, isola2014what, isola2011makes,khosla2015understanding}. 
And, interestingly, despite people being remarkably consistent in what they remembered and forgot, their subjective judgements about what was memorable versus forgettable were inaccurate~\cite{isola2014what}. 
Thus, memorability was found to be intrinsic to the images studied, largely independent of the observers and their opinions, and yet not easily described by previously-studied high- or low-level features \cite{rust_understanding_image_memorability}. 

These discoveries sparked a line of research dedicated to building computational models to understand and predict image memorability.
The memorability games first introduced by Isola et al.~\cite{isola2011makes} became the gold standard for collecting objective measurements of human memorability~\cite{bainbridge2013intrinsic,borkin2013makes,bylinskii2015intrinsic,khosla2015understanding} and the basis for memorability datasets that fueled further work on memorability prediction.
% and were widely adopted in later studies 
% were thus adopted with minimal changes in follow-up studies~\cite{bainbridge2013intrinsic,borkin2013makes,bylinskii2015intrinsic,khosla2015understanding} 
% to collect these objective measurements of human perception and cognition, and study them more closely. 
% A line of research was then sparked, dedicated to building computational models to understand and predict image memorability, 
Driven by the promise of automatic applications for memory manipulation, assistive devices, and more effective visuals, computational models of memorability became more complex and extended to cover many types of stimuli. 

This chapter will provide an overview of these computational efforts. In Section~\ref{sec:datasets}, we start our discussion with the datasets that power the computational models of memorability. We mention considerations for data curation and provide a list and descriptions of the image memorability datasets available to date, how they were sourced, and which additional properties they contain. Section~\ref{sec:models} provides an overview of different model designs, from support vector machines to different types of neural networks (including CNNs, RNNs, and GANs). We discuss model design considerations, including interpretability, and provide a curated list of the top-performing published models. In Section~\ref{sec:features}, we cover our current understanding about which features of images and videos make them more or less memorable, from low-level pixel features to semantic and contextual features, including objects, saliency, motion, and emotion. We wrap up with a discussion of applications in Section~\ref{sec:applications}, future directions for research in Section~\ref{sec:future}, and our proposed unifying explanation for the ``magic sauce'' of memorability in Section~\ref{sec:conclusion}.

%the model designs that have proven successful, and our current understanding about which features of images make them more or less memorable, concluding with a preview of possible applications. 

% ####################################################

\section{Datasets: from visual content to scores}\label{sec:datasets}

Data is at the heart of most computational models. A good dataset can make a simple model shine; a poor dataset will undercut even the most apt modeling decisions. We begin this chapter with a discussion of the factors that are important for collecting a memorability dataset, followed by a brief overview of existing datasets. Memorability scores for the stimuli in all these datasets were collected using a variation on the memory game protocol from the seminal memorability paper by Isola et al.~\cite{isola2011makes}. %A ``how-to" guide for setting up such memory games is provided in the previous chapter; 
This section will focus on the stimuli used for the memorability games.

Designing datasets for memorability studies requires careful data curation, so that neither the insights obtained from the analyses nor the models trained on this data are biased due to confounding variables. The following properties should be considered:
%Memorability datasets should be collected with the following considerations:
\begin{itemize}
    \item \textbf{Diversity}. Collecting stimuli with good variability along the relevant feature axes will facilitate spread in the human memorability scores and allow models to learn a more robust signal from the data~\cite{khosla2015understanding}. Even within a single scene category, images can vary in terms of the objects contained, the viewing angle, the amount of light, etc.~\cite{bylinskii2015intrinsic,goetschalckx2019memcat}. 
    \item \textbf{Quantity}. Having a large number of stimuli will have two benefits: (i) the effects of confounding variables have a larger chance of being washed out, and (ii)~larger, more powerful models can be trained when more data is available. The best performing models today are particularly data-hungry neural networks~\cite{fajtl2018amnet,khosla2015understanding,perera2019image}. Having a sufficient number of participant responses per stimuli is also important, as having too few participants can produce an artificially low split-half consistency value.
    The Spearman-Brown formula \cite{Spearman_1910, Brown_1910, goetschalckx2018longer} can be used to calculate an appropriate number of responses to collect in order to reach a stable value for split-half consistency.
    \item \textbf{Balance}. It is important to decide up front whether to explicitly balance the data by having similar numbers of exemplars per category~\cite{bylinskii2015intrinsic,goetschalckx2019memcat} or to sample according to some natural distribution (e.g., sampling photos of faces according to names from the U.S. census~\cite{bainbridge2012establishing}). If a dataset has cluttered indoor scenes, one can consider the addition of cluttered and uncluttered outdoor as well as uncluttered indoor scenes~\cite{bylinskii2015intrinsic}. For specialized datasets like faces, consider genders and races~\cite{bainbridge2012establishing}; for graphic designs, figures, or visualizations, consider publication sources and design categories~\cite{borkin2013makes}. Building off prior work can help balance for visual features and semantic content that has been previously found to drive memorability (e.g., the presence of faces, emotional content, zoomed-in objects, etc.; see Section~\ref{sec:features}).
\end{itemize}

Further, there are a number of considerations at play when assembling the stimuli for use in memorability studies. They are outlined below.
%\textbf{Sourcing stimuli for datasets.} 

\textbf{Permissions and appropriateness}. Stimuli used in memorability datasets are often drawn from previously-curated open-source datasets used in computer vision, including  
the SUN scenes dataset \cite{isola2011understanding,xiao2010sun}, Aesthetic Visual Analysis (AVA) \cite{khosla2015understanding,murray2012ava}, Abnormal Objects \cite{khosla2015understanding,saleh2013object}, or Moments in Time \cite{monfort2019moments, newman2020memento}.
As with any image or video dataset, it is important to take care that you have permission to use and share the stimuli before publishing your data and to filter for quality and appropriateness.

\textbf{Filtering to avoid confounds}. Once sourced, some preprocessing of the data may be required, in particular to guarantee that irrelevant factors (e.g., the size, aspect ratio, image quality, or speed - in the case of videos) are held constant, to avoid confounding the results. Some of the processing, including cropping image size~\cite{bylinskii2015intrinsic,isola2011makes} can be done automatically, while the rest may require manual curation. This type of curation can be amenable to crowdsourcing tasks (e.g., removing stimuli with undesirable properties like watermarks, special effects, etc.~\cite{newman2020memento}).

\textbf{Controlling for familiarity}. Finally, the memorability studies described in this work assume that participants are seeing the stimuli for the first time. To avoid confounding familiarity effects, it may be necessary to filter the dataset ahead of time to remove landmarks, faces, artwork, and other potentially recognizable content.

\subsection{Consistency across participants}
\begin{figure}
\centering
\includegraphics[width=0.8\linewidth]{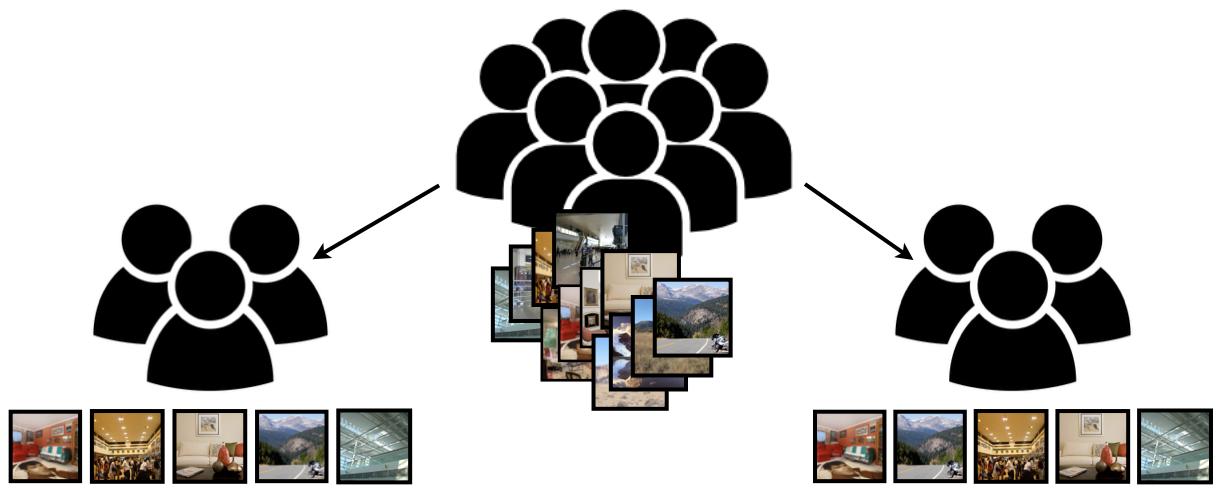}
\caption{Consistency is evaluated by randomly splitting the participants in a memorability experiment in half, computing memorability scores based on the data from each set of participants separately, and ranking the dataset images according to those scores. The correlation between the sets of rankings is reported. This procedure is repeated for multiple splits, and the results are averaged to produce the final consistency score ($\rho$).}
\label{fig:consistency}
\end{figure}

%Despite differences across individuals and populations, results show 
Many studies have confirmed that the memorability of %different types of 
visual stimuli - be they natural scenes, portraits, visualizations, objects, or actions - is remarkably consistent across viewers (Figure~\ref{fig:consistency}).
This level of consistency between participants is itself roughly consistent across studies with different participant groups and stimulus sets, as shown in Table~\ref{tab:datasets}. 
Consistency is measured as the Spearman rank correlation between memorability rankings produced by different groups of participants, and falls in the range 0.68-0.83 for image datasets and 0.57-0.73 for video datasets. Note that split-half consistency tends to increase with number of participants and levels off with sufficient data points.
Early studies~\cite{isola2011understanding,isola2014what,isola2011makes} showed that indoor images with people are consistently more memorable than natural scene images, but this observation did not fully account for inter-observer consistency.
% the consistency observed across people. 
Later, studies run within-category, where only indoor images of a single scene category were shown (e.g., kitchens), continued to exhibit similarly high inter-observer consistency scores~\cite{bylinskii2015intrinsic}. Analogous levels of consistency were observed once scene and object classes were more carefully controlled for~\cite{dubey2015makes,goetschalckx2019memcat,9025769}. Specialized image sets including face images~\cite{bainbridge2013intrinsic} and information visualizations~\cite{borkin2013makes}, as well as video datasets \cite{cohendet2019videomem, newman2020memento}, maintained high inter-observer consistency, once again validating the intrinsic nature of memorability. %Table~\ref{tab:datasets} summarizes the image and video memorability datasets published to date. 

High inter-observer consistency means that memorability scores are eligible for automatic prediction. The rest of this section will go into greater detail about how the stimuli for these memorability studies were collected and the additional annotations that they contain, which will be useful for the computational analyses and models in the following sections.

\begin{table}[!t]
\caption{Image and video memorability datasets.}
\label{tab:datasets}       % Give a unique label
%
% Follow this input for your own table layout
%
\begin{tabular}{p{3.0cm}p{1.1cm}p{1.1cm}p{6.1cm}}
\hline\noalign{\smallskip}
Dataset & Num. Stimuli & $\rho^*$ & Type of stimuli  \\
\noalign{\smallskip}\svhline\noalign{\smallskip}
SUN (2011) \cite{isola2011makes} & 2,222 & 0.75 & Mixed photographs \\
Face Mem. (2013) \cite{bainbridge2013intrinsic} & 2,222 & 0.68 & US adult faces \\
MASSVIS (2013) \cite{borkin2013makes} & 393 & 0.83 & Data visualizations and infographics \\
FIGRIM (2015) \cite{bylinskii2015intrinsic} & 1,754 & 0.74 & Scenes from 21 indoor and outdoor categories \\
Object Mem. (2015) \cite{dubey2015makes} & 850 & 0.76 & Scenes with object segmentations \\
LaMem (2015) \cite{khosla2015understanding} & 60,000  & 0.68  & Diverse images from other computer vision datasets\\
MemCat (2019) \cite{goetschalckx2019memcat} & 10,000 & 0.78 & 5 broad image categories (+ sub-categories) \\ 
VISCHEMA (2019) \cite{akagunduz2019defining} & 800 & - & Subset of FIGRIM with additional annotations \\ 
LNSIM (2020)~\cite{9025769} & 2,632 & 0.78 & Natural outdoor scenes without salient objects \\ \hline
Movie Mem. (2018)~\cite{cohendet2018annotating} & 660 & 0.57 & Video clips from previously-seen films \\
VideoMem (2019)~\cite{cohendet2019videomem} & 10,000 & 0.62 & Seven-second video clips \\
Memento10k (2020)~\cite{newman2020memento} & 10,000 & 0.73 & Three-second, dynamic video clips \\

\noalign{\smallskip}\hline\noalign{\smallskip}
\end{tabular}
$^*$ Consistency refers to inter-observer split-half consistency in the hit rates (HR), measured with Spearman correlation, and reported as $-1 \leq \rho \leq 1$. The number of stimuli listed in the table above corresponds to those with memorability scores (on which consistency scores could be calculated). Many of the datasets above also contain a larger number of curated stimuli without memorability scores, used as fillers in the memorability games. Other studies~\cite{khosla2015understanding,newman2020memento} used the same stimuli as targets and fillers for different participants, thereby not requiring separate sets of stimuli.
\end{table}

\subsection{Natural scenes}

The original memorability dataset by Isola et al.~\cite{isola2011makes} contains a random sampling of scene categories from the SUN dataset~\cite{xiao2010sun}, of which 2,222 images have memorability scores. 
The images were resized and cropped to the same size, so that neither size nor aspect ratio acted as confounding variables.
They also come with human-annotated object segmentation labels 
%into object regions with labels given by a human annotator using the LabelMe interface
\cite{russell2008labelme}, which were originally used to correlate memorability with various object classes. A follow-up work by Isola et al.~\cite{isola2014what} also collected aesthetic and interestingness judgements for each of these images, as well as subjective judgements as to whether humans consider them memorable. 

The FIGRIM dataset~\cite{bylinskii2015intrinsic} sampled scene categories in a more targeted way. It represents 21 indoor and outdoor SUN scene categories, each of which has sufficient exemplars at sufficient resolution (at least 700x700 pixels; as above, images were preprocessed to a consistent size before the experiment). 
%Out of the resulting dataset of 9,428 images, 
Memorability scores were collected for 1,754 target images. An additional 7,674 filler images are available.

LNSIM~\cite{9025769} was a dataset intended to capture the memorability of scenes without the confounding effects of salient and memorable objects. Towards this goal,
%, and mimicking the data collection procedure of LaMem~\cite{khosla2015understanding}, 
the authors collected images from MIR Flickr~\cite{huiskes2008mir}, AVA~\cite{murray2012ava}, affective images~\cite{machajdik2010affective}, the image saliency datasets MIT1003~\cite{judd2009learning} and NUSEF~\cite{ramanathan2010eye}, and SUN~\cite{xiao2010sun}.
They further filtered % had a group of human annotators hand curate 
the images to ``only be composed of outdoor natural scenes not having any human, animal and man-made object''. 
Memorability scores for 2,632 images were measured based on an average of 80 observers per image and calculated using the same procedure as the LaMem dataset~\cite{khosla2015understanding}.
% Aside from memorability scores captured from an average of 80 observers per image, each of the 2,632 images in LNSIM 
The images come hand-tagged with scene category labels from 71 scene categories obtained from WordNet~\cite{miller1995wordnet}.

\subsection{Diverse photographs}

\begin{figure}
\centering
\includegraphics[width=1\linewidth]{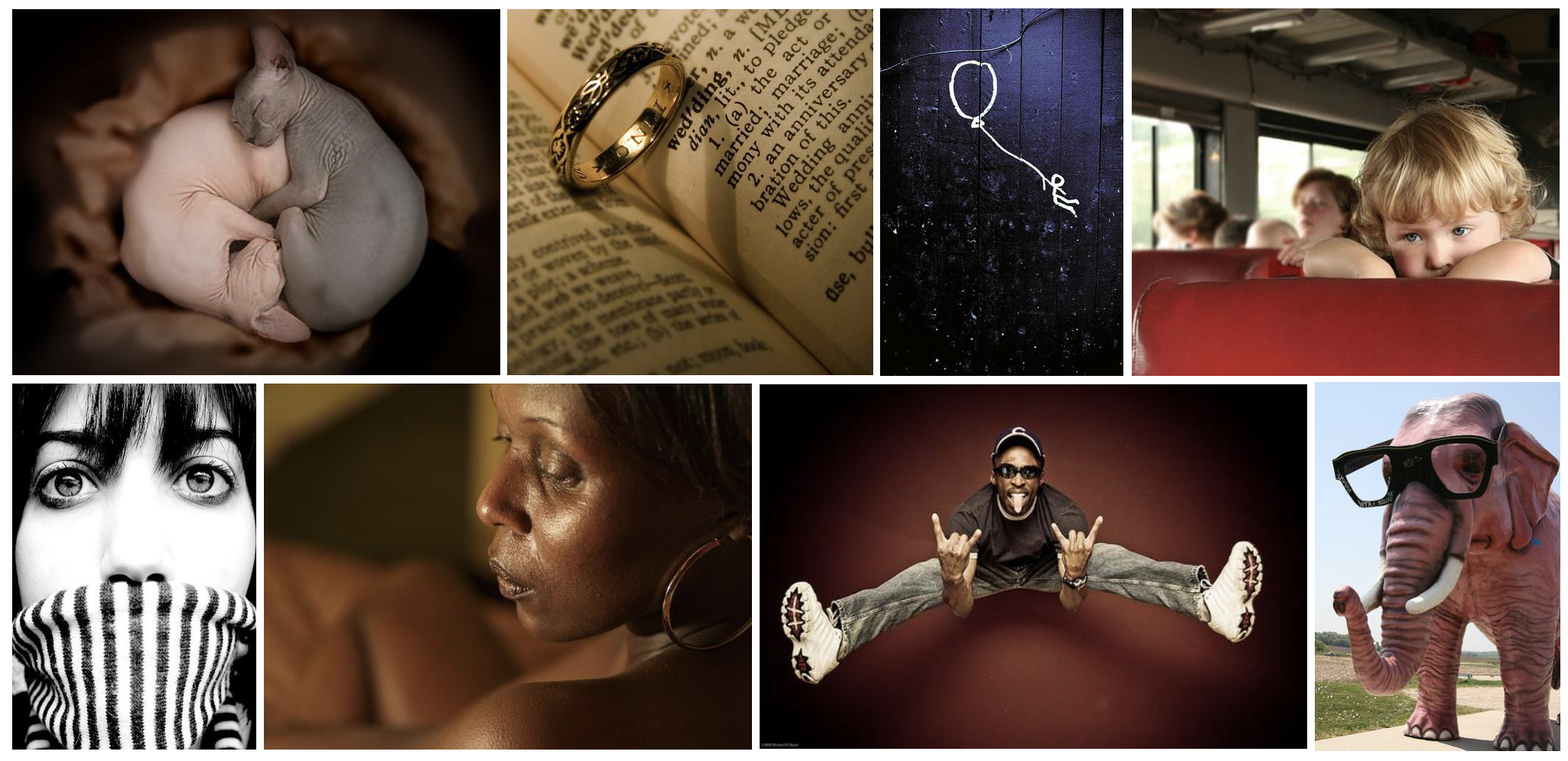}
\caption{LaMem~\cite{khosla2015understanding} contains 60k images sampled from different computer vision datasets varying in the types of scenes and objects portrayed, as well as in aesthetic and affective value. All images are provided with collected memorability scores.}
\label{fig:lamem}
\end{figure}

LaMem~\cite{khosla2015understanding} is the largest and most varied image memorability dataset collected to date (Figure~\ref{fig:lamem}). Its size--60k images with memorability scores--makes it particularly effective for training machine learning models. Its images come from MIR Flickr~\cite{huiskes2008mir}, AVA~\cite{murray2012ava}, affective images~\cite{machajdik2010affective}, the image saliency datasets MIT1003~\cite{judd2009learning} and NUSEF~\cite{ramanathan2010eye}, SUN~\cite{xiao2010sun}, an image popularity dataset~\cite{khosla2014makes}, the Abnormal Objects dataset~\cite{saleh2013object}, and PASCAL~\cite{farhadi2009describing}. Thus, these images contain a variety of object and scene types and vary in aesthetic and affective value. 
Since subsets of the LaMem images 
% These datasets 
contain additional labels, Khosla et al.~\cite{khosla2015understanding} were able to analyze how a variety of factors correlated with memorability, 
%that could be compared to memorability, 
including popularity scores~\cite{khosla2014makes}, eye fixations~\cite{judd2009learning}, emotions~\cite{machajdik2010affective}, and aesthetic scores~\cite{murray2012ava}. % These analyses can be found in Khosla et al.~\cite{khosla2015understanding}.

The MemCat image set has a category-based structure, with images belonging to five broad semantic categories (animal, food, landscape, sports, and vehicle), each with at least 20 sub-categories (e.g., animal: bear, duck; food: burrito, salad; landscape: desert, lake; sports: baseball, surfing; vehicle: airplane, train). 
%making for a large and varied set. 
% Those specific five categories were chosen in part because they were expected to differ considerably in memorability. 
While the broad category labels explained roughly 40\% of the variance in the MemCat scores, images of the same category still differed consistently in memorability. With 2K images per category, one of MemCat's intended uses is to promote further research into within-category memorability variance. Images were taken from ImageNet \cite{deng2009imagenet}, COCO \cite{lin2014coco}, SUN \cite{xiao2010sun}, and the Open Images Dataset \cite{kuznetsova2018open}. The source data sets offer additional annotations such as bounding boxes or segmentation maps. 
% Finally, the set was composed with caution regarding 
Finally, the authors took care to account for potentially confounding influences including the presence of people in non-people categories, large readable text, oddities, etc.
% of the presence of people in non-people categories (all but sports), large readable text, oddities, etc.

Dubey et al.~\cite{dubey2015makes} sampled images from the PASCAL-S dataset~\cite{li2014secrets}, a fully segmented subset of PASCAL VOC 2010~\cite{everingham2011pascal}, and cleaned up the segmentations to only contain clearly-visible, complete, and nameable objects. The resulting dataset contains 850 images with a total of 3,412 segmented objects ($\sim$4 objects per image). Rather than capturing the memorability of whole images using the set-up from prior work~\cite{isola2011makes}, whereby images presented for encoding and recognition are interspersed, Dubey et al. first showed participants whole images in an encoding stage, followed by individual objects (extracted from images) in a recognition stage to capture the memorability of individual objects~\cite{dubey2015makes}. As shown in~\cite{goetschalckx2018longer}, memorability rankings collected with an interspersed paradigm versus a paradigm with separate stages are highly correlated.

\subsection{Specialized image collections}

The original studies on scene memorability~\cite{isola2011understanding,isola2011makes} sparked research on specialized stimuli collections, including faces and information visualizations. The 10k US Adult Faces Database~\cite{bainbridge2012establishing,bainbridge2013intrinsic} contains 10,168 face photographs obtained by sampling the US Census for first and last names and downloading color face photographs available via Google Image Search, manually filtered to exclude celebrities, children, and low-quality or unusual images. The resulting database follows the gender, age, and race distributions of the adult US population. For 2,222 face photographs, memorability scores were collected, along with another 20 facial and personality traits (using a crowdsourcing survey). Bainbridge et al.~\cite{bainbridge2013intrinsic} then reported which of these traits were correlated with face memorability.

Borkin et al.~\cite{borkin2013makes} collected a total of 2,070 data visualizations and infographics from different publication sources, which were then hand-tagged with taxonomic category (bar graph, table, diagram, map, etc.). Memorability scores were collected using the standard study setup for 393 of these visualizations, which were also hand-tagged with multiple attributes (clutter, colorfulness, etc.). Follow-up work~\cite{borkin2015beyond} segmented the visualizations into components (using LabelMe~\cite{russell2008labelme}), and obtained eye movements as well as captions about the visualizations from memory. Additional memorability scores were collected using a modified study setup and longer viewing durations (10 seconds/image), to account for the increased information content in visualizations compared to natural images.

\subsection{Videos}
Large-scale video memorability datasets have emerged to spur research into memorability of dynamic stimuli. The currently available datasets have focused on short video clips (<10 seconds) that are treated as discrete stimuli without audio for the purposes of continuous-recognition memorability experiments.

VideoMem \cite{cohendet2019videomem} consists of 10,000 seven-second video clips taken from raw footage intended for reprocessing into advertisements or television shows. 
As such, the videos consist primarily of staged, professionally-shot scenes.
VideoMem features a variety of content including people, animals, inanimate objects, and nature scenes.
It contains memorability scores at two different delays: a short delay (a few minutes after viewing) and a longer delay (a couple days after viewing) to enable study into how memorability rankings may or may not reorder over time.

Memento10k \cite{newman2020memento} contains three-second ``in-the-wild'' video clips.
These videos were originally posted to media sharing sites like YouTube and Flickr. 
They were then manually filtered to remove objectionable content or artificial artifacts like captions, cartoons or post-processing.
Like VideoMem, Memento10k contains varied semantic content, encompassing people, animals, objects, and landscapes; however, because the videos were shot by laypeople, Memento10k encompasses more variability in levels of motion and video quality.
For each Memento video, five human-written captions were collected describing the contents of the clip.
% Memento10k contains additional annotations in the form of human-written captions.

Earlier work on computational video memorability attempted to identify the features that contributed to memorability, but often relied on smaller datasets and paradigms that were more challenging to scale than the popular continuous-recognition experiments.
Cohendet et al.~\cite{cohendet2018annotating} collected the Movie Memorability Dataset, which measures long-term memorability over weeks or years using clips from popular movies. 
Their experimental setup differs somewhat from the traditional memory game experiment in that the first viewing of the stimuli took place before the start of the experiment, and viewing delay was estimated by asking participants when they had watched the movie in question.
The authors found that semantic features derived from video captions were most correlated with the memorability scores of these clips.
Han et al.~\cite{han2015fmri} leveraged fMRI data obtained from participants viewing videos in order to produce a computational model that aligns audiovisual features with brain data. % in order to improve prediction.
Shekhar et al.~\cite{shekhar2017show} derived memorability ratings for 100 videos using response time on a verbal recall task, and explored the contribution of deep, semantic, saliency, spatio-temporal, and color features to predict these scores.

% ####################################################

\section{Models: from pixels to features}\label{sec:models}

%\TODO{Papers left to go over and add content from: \cite{zarezadeh2017image,borkin2013makes,bylinskii2015intrinsic,bainbridge2013intrinsic,khosla2013modifying,jing2016predicting}}

%\TODO{Lore: I added/suggested stuff from \cite{kim2013relative, mancas2013memorability, 9025769, akagunduz2019defining, celikkale2015predicting, khosla2012image}}

In the last section we discussed 
%Analyses performed on the memorability scores collected by past studies validated 
that memorability can be treated as an image-computable measure, largely independent of the observer. In this section we review the range of computational models that have been applied to predicting image and video memorability, from the more traditional machine learning models to state-of-the-art deep neural networks. In all cases, these models have been trained on memorability datasets and evaluated on either a held-out portion of the same dataset or on a separate memorability dataset altogether. A summary of the different image memorability models is provided in Table~\ref{tab:immodels}, with the reported prediction scores taken directly out of the published papers (whenever available). Similarly, Table~\ref{tab:vidmodels} contains the prediction scores of video memorability models.

\begin{table}
\centering
\caption{Model performance on image memorability prediction, as reported in past papers. This is the Spearman correlation between the predicted and measured memorability scores, with human upper bound defined as inter-observer consistency.}
\label{tab:immodels}
\begin{tabular}{llll}
\hline\noalign{\smallskip}
\multicolumn{1}{c}{Model} & \multicolumn{3}{c}{Dataset} \\
 & LaMem~\cite{khosla2015understanding} & SUN~\cite{isola2011makes} & Figrim~\cite{bylinskii2015intrinsic} \\
\noalign{\smallskip}\svhline\noalign{\smallskip}
Human (upper bound) & 0.68 & 0.75 & 0.74 \\ \hline
MemBoost~\cite{perera2019image} & 0.67 & 0.66 & 0.57 \\
%AMNet~\cite{fajtl2018amnet} & 0.677 & 0.649 & - \\
AMNet~\cite{fajtl2018amnet} & 0.68 & 0.65 & - \\
MemNet~\cite{khosla2015understanding} & 0.64 & 0.63 & - \\
%CNN-MTLES~\cite{jing2016predicting} & 0.5025 & - & -\\
CNN-MTLES~\cite{jing2016predicting} & 0.50 & - & -\\
%MemoNet~\cite{baveye2016deep} & - & 0.636 & -\\
MemoNet~\cite{baveye2016deep} & - & 0.64 & -\\
%Hybrid-CNN+SVR~\cite{zarezadeh2017image} & - & 0.6202 & -\\
Hybrid-CNN+SVR~\cite{zarezadeh2017image} & - & 0.62 & -\\
%Mancas \& Le Meur~\cite{mancas2013memorability} & - & 0.479 & -\\
Mancas \& Le Meur~\cite{mancas2013memorability} & - & 0.48 & -\\
%Isola~\cite{isola2011makes} & - & 0.462 & - \\
Isola~\cite{isola2011makes} & - & 0.46 & - \\
\end{tabular}
\end{table}
% Lore: I double checked an all I could find is that MemoNet was also tested on a small set of 150 IAPS images that they quantified themselves, but with no details on how many responses per image, consistency, etc. So not worth mentioning, I feel

\begin{table}
\centering
\caption{Model performance on video memorability prediction.}
\label{tab:vidmodels}
\begin{tabular}{lll}
\hline\noalign{\smallskip}
\multicolumn{1}{c}{Model} & \multicolumn{2}{c}{Dataset} \\
 & VideoMem (val)~\cite{cohendet2019videomem} & Memento10k (test)~\cite{newman2020memento} \\
\noalign{\smallskip}\svhline\noalign{\smallskip}
Human (upper bound) & 0.616 & 0.730 \\ \hline
SemanticMemNet~\cite{newman2020memento} & 0.556 & 0.663 \\
VideoMem-Semantic~\cite{cohendet2019videomem} & 0.503 & 0.552 \\
MemNet (frames baseline) \cite{khosla2015understanding} & 0.425 & 0.485 \\
\end{tabular}
\end{table}

\subsection{Support Vector Machines}

Before neural networks became the tool of choice for many computer vision prediction tasks, a common machine learning approach was to extract low-level feature descriptors like GIST~\cite{oliva2001modeling}, HOG~\cite{dalal2005histograms} or SIFT~\cite{lowe2004distinctive} from images, often at multiple spatial scales~\cite{lazebnik2006beyond}, and then to train a Support Vector Machine (SVM) to map from the assembled feature vectors to the target labels.
%, using splits of training/test data. 
For problems with real-valued labels (like memorability scores), Support Vector Regression (SVR) was used instead~\cite{drucker1996support,fan2008liblinear}. 
In the case of memorability, 
%among the low-level features, 
it was found that HOG2x2~\cite{khosla2012memorability,wang2010locality} was one of the most predictive automatic image features when used with a linear SVR machine~\cite{isola2014what,khosla2013modifying,khosla2012memorability}.  
%Mancas and LeMeur~\cite{mancas2013memorability} later showed that attention-related features can replace some of the low level features

\textbf{Differential weighting of features}: Instead of extracting image features from the image as a whole, another approach is to differentially weight features obtained from different regions of an image, if there is good reason to suppose that certain image regions should contribute more to the prediction. For instance, multiple studies have reported that not all image regions are remembered equally well~\cite{akagunduz2019defining,khosla2012image}. Akagunduz et al.~\cite{akagunduz2019defining} explicitly ask participants in a memory task to indicate which regions made them recognize an image. Pooling this information across participants results in consistent ground truth maps which the authors call Visual Memory Schemas (VMS). They report that spatially pooling and weighting image features (e.g., GIST, SIFT, HOG) by the VMS not only yields better SVR predictions than without doing spatial pooling, but also outperforms spatial pooling and weighting with eye-fixation maps or saliency maps. In later work, the authors also propose a way to predict the VMS of an image itself \cite{kyledavidson2020predicting, kyledavidson2020generating}. Inspired by the observation that visual attention is highly related to memory~\cite{hollingworth2002accurate,hollingworth2001see,wolfe2007visual}, some work has also looked at spatially pooling visual features based on computational saliency maps~\cite{celikkale2015predicting}.  

\begin{svgraybox}
\textbf{What is SVM and SVR?} \\ \\
A Support Vector Machine (SVM) is a machine learning model that classifies data points 
%into different categories 
by finding a hyperplane in feature space that separates the training data into different categories.
%into their respective classes.
% , using the provided labels. 
% The decision boundaries are defined by the points that are hardest to classify - they are called the support vectors, hence the name of the model. 
At training time, the SVM takes in labeled training points and finds a decision boundary that correctly separates the data with the biggest possible margin, or distance between the data and the correct side of the boundary.
% learns to maximize the distance between the data points and the decision boundary.
At test time, prediction involves determining which side of the decision boundary the test points are on 
% (by comparing them to the support vectors) 
in order to assign them a label. 
Support Vector Regression (SVR) extends the formulation of SVMs to regression problems with real-valued labels, instead of class labels, by finding the hyperplane that best fits the greatest number of training points. At test time, prediction involves finding the point on the hyperplane corresponding to the test data point.
\end{svgraybox}

\subsection{Convolutional Neural Networks}

\textbf{Deep features}: Deep representations learned by convolutional neural networks (CNNs) have proven to be the most successful computational features at approximating human memorability~\cite{dubey2015makes,khosla2015understanding,shekhar2017show}. 
These features are extracted from a layer (most commonly the penultimate one) of a pre-trained neural network, which has often been trained for the task of ImageNet~\cite{russakovsky2015imagenet} classification.
Later layers of neural networks are known to capture image semantics, in the form of distributions of objects over the image~\cite{bylinskii2015intrinsic,sharif2014cnn}. For instance, Dubey et al.~\cite{dubey2015makes} used features extracted from the AlexNet CNN~\cite{krizhevsky2012alexnet} trained on ImageNet 
% (using the Caffe machine learning library~\cite{jia2014caffe}) 
to predict object memorability. 

Apart from using CNNs for prediction, there is a second way in which CNNs can provide insight into memorability: as a model for the way memorability emerges from neural processing in the brain \cite{rust_understanding_image_memorability, Jaegle2019}. 
In particular, Jaegle et al.~\cite{Jaegle2019} found that in the later layers of a CNN trained for image classification, the response magnitude of nodes in the network correlate with memorability. This is consistent with the authors' observation that population response magnitude in the IT cortex of the (monkey) brain predicts memorability. 

\textbf{Transfer learning}: Transfer learning is a common strategy that can increase model performance when the target task--in our case, memorability prediction--has a relatively small dataset.
The simplest form of transfer learning involves pre-training a network on a task for which a large dataset is available and then fine-tuning some subset of the network's layers on the target task with the target dataset. 
MemNet~\cite{khosla2015understanding}, for example, is based on the AlexNet architecture~\cite{jia2014caffe,krizhevsky2012alexnet} that was pre-trained for a classification task on a 3.6-million image dataset consisting of a combination of ImageNet~\cite{russakovsky2015imagenet} and Places~\cite{zhou2014learning}. 
Only then was it fine-tuned on the 60k-image LaMem dataset to predict real-valued memorability scores~\cite{khosla2015understanding}.
% MemNet~\cite{khosla2015understanding}, for example, is based on the AlexNet architecture~\cite{krizhevsky2012alexnet,jia2014caffe} trained on object and place recognition (using the ILSVRC dataset~\cite{russakovsky2015imagenet} and Places~\cite{zhou2014learning} dataset, containing $\sim$ 3.6 million images in total) to predict object/scene classes as Hybrid-CNN~\cite{zhou2014learning}, and then fine-tuned on the 60K image LaMem dataset~\cite{khosla2015understanding} to predict a real-valued memorability score, using Euclidean loss. 
Similarly, MemoNet~\cite{baveye2016deep} is a fine-tuned GoogleNet model~\cite{szegedy2015going}. 
AMNet~\cite{fajtl2018amnet} includes as a backbone a fine-tuned ResNet model~\cite{he2016deep}. Perera et al.~\cite{perera2019image} found that fine-tuning the last (regression) layer performed better than fine-tuning or retraining the whole network, because this approach was less likely to suffer from over-fitting to the training set. 
Video models have turned to even more complex pre-training regimes. SemanticMemNet~\cite{newman2020memento}, which predicts video memorability, contains an image-based stream that was pre-trained on ImageNet and LaMem, as well as video and optical flow streams pre-trained on ImageNet and Kinetics~\cite{carreira2017kinetics}. The semantic embedding model from~\cite{cohendet2019videomem} used as its base a video captioning model~\cite{engilberge2019sodeep} that was also pre-trained on LaMem.

\textbf{Size matters}: Perera et al.~\cite{perera2019image} experimentally confirmed that the more powerful classification networks performed better on memorability prediction (i.e., AlexNet performing worst, and ResNet152~\cite{he2016deep} best).
%In general, deeper networks are often found to perform better, provided proper training procedures are used to prevent over-fitting on the training data. 
As CNN architectures continue to improve, we may find that more powerful backbones contribute to improved performance on memorability prediction, particularly in the relatively new domain of video prediction.
% We can thus expect that as CNN architectures continue to improve, so will the corresponding memorability prediction networks built upon them. 
However, the amount of data available is an important factor as well. As a case in point, Khosla et al.~\cite{khosla2015understanding} found that when fine-tuning on the larger LaMem dataset, the resulting network performed better on both LaMem and SUN datasets, even compared to a network trained and tested on the SUN dataset. This is because having too little data with which to train or fine-tune can result in model over-fitting, which reduces the generalization ability of the network on held-out data, regardless of which dataset it comes from. 

\begin{svgraybox}
\textbf{What is a CNN?} \\ \\
A Convolutional Neural Network (CNN) is a machine learning model that learns a very complex function to map an input--often an image or video--to a desired output, given a large dataset. 
It works by applying a sequence of linear and non-linear operations to the input.
In a CNN, the linear operations are most frequently convolutions, thus giving the model its name. 
% , which are grouped together into layers.  
%The operations are stacked as layers of a network. 
% Groups of operations are referred to as layers.
Each operation is defined by a set of parameters or weights, which must be learned from the training data. 
The learning process involves optimizing the output of the network in order to minimize some loss function (for example, a classification or regression loss) using gradient descent.
% The learning is an optimization problem that depends on a loss function (e.g., classification or regression loss). 
Once trained, the sequence of operations with learned parameters can be applied to a new input to produce the final prediction.
% a CNN can be applied to a test input, which involves a quick forward pass through the linear and non-linear operations (with now fixed parameters) to produce a final prediction. 
The sheer number of degrees of freedom in these models (due to the number of operations and parameters), as well as the sophisticated optimization algorithms used for training them, make for powerful models that are popular for a large range of tasks.
\end{svgraybox}

\subsection{Recurrent Neural Networks}
Rather than processing an image all at once as in a standard CNN architecture, Recurrent Neural Networks (RNNs) allow the model to parse an image in pieces, aggregating evidence before making a prediction. Fajtl et al.~\cite{fajtl2018amnet} showed gains in performance from using an RNN-based architecture with soft attention in order to make three passes over an image, each time focusing on a different set of image regions, before predicting the final memorability score. 
Using a larger and more complex model than the CNN-based MemNet~\cite{khosla2015understanding} gives this model greater predictive power. 
Another advantage of this model is that the soft attention can be visualized as heatmaps over the image regions attended to at each of the three passes. This is useful for determining what evidence the model uses for its predictions, and which parts of the image are most informative for determining the memorability score. 

Newman et al.~\cite{newman2020memento} also explored using RNNs to improve memorability prediction on videos. 
% RNNs, which are popular models for sequential inputs, are a natural choice for video inputs.
LSTMs \cite{lstm} are a type of RNN that are frequently used for processing or generating natural language. 
SemanticMemNet uses an LSTM \cite{lstm} to predict verbal captions for a video.
It uses the additional supervision provided by the language descriptions in Memento10k \cite{newman2018effects} to encourage learning a feature space that explicitly encodes semantic features. 
% Newman et al.~\cite{newman2020memento} 
The authors also experimented with using an RNN to predict raw hit rates at different viewing delays.

\begin{svgraybox}
\textbf{What is an RNN?} \\ \\
Recurrent Neural Networks (RNNs) are neural networks with ``loops" in them, where the prior state of the network is fed as input to future states, allowing previously gathered information to persist. They are popular for sequential tasks, most commonly language-related tasks like translation and image captioning. However, they are also used in cases where processing an input in pieces makes sense. For instance, when combined with soft attention, they can process an image not all at once, but in a sequence of ``glimpses", where each ``glimpse" focuses the network on a particular region of an image.  One of the most common types of RNNs is an LSTM (Long short-term memory)~\cite{lstm}, popular for its ability to capture long-term dependencies via a set of computational control gates specifically designed to let some information flow through unchanged, and other information to be attenuated. 
\end{svgraybox}

\subsection{Generative Adversarial Networks}
\begin{figure}
\centering
\includegraphics[width=1\linewidth]{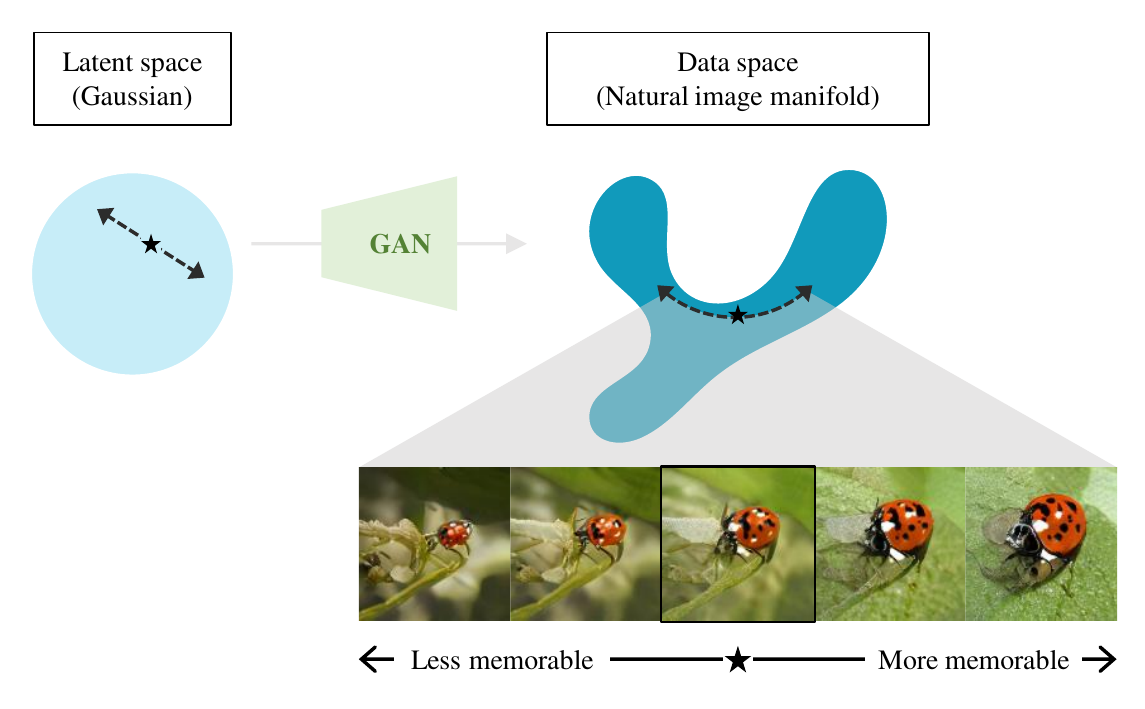}
\caption{A GAN can be used to transform samples drawn from some latent space (for instance, samples of Gaussian noise) into samples lying in the data space that the model has been trained on - e.g., approximating natural images. Moving along various latent directions, the model can generate novel outputs, that have not otherwise been previously seen. See Jahanian et al.~\cite{gansteerability} for a discussion about the steerability of GAN models.}
\label{fig:ganexample}
\end{figure}

Another technique for understanding image memorability, beyond predicting memorability scores, is to directly visualize the qualities that make an image memorable. 
Goetschalckx et al.~\cite{goetschalckx2019ganalyze}, for instance,
%dug deeper for a finer-grained understanding of image memorability. They 
re-purposed a Generative Adversarial Network (GAN) \cite{goodfellow2014generative} to explore image perturbations that either increase or decrease the memorability of images (Figure~\ref{fig:ganexample}). 
They found directions in latent space that best correlated with changes in image memorability, as measured by the MemNet model \cite{khosla2015understanding}. 
By automatically visualizing increasingly memorable images, they could observe how the GAN modifies the composition of the image and the objects within.

Rather than using a pre-trained GAN, related work \cite{kyledavidson2020generating} trained a GAN from scratch and had it accept an additional input value (M) representing a desired memorability score for the output image. The memorability of the output was measured by an auxiliary model based on the aforementioned Visual Memory Schemas \cite{akagunduz2019defining}. By varying M while keeping the latent vector (i.e., the standard GAN input) constant, this framework also produces visualizations of images gradually increasing in memorability. 

\begin{svgraybox}
\textbf{What is a GAN?} \\ \\
Generative Adversarial Networks (GANs) \cite{goodfellow2014generative} are most commonly used to synthesize fake images that look realistic.
They work by generating convincing, novel samples from a very complex, high-dimensional data distribution--in this case, the distribution of natural images.
% aim to model the full, joint distribution over high-dimensional, complex sensory data, in this context the distribution of natural images. A GAN can then sample from this distribution to synthesize new, artificial images that still look realistic. 
They are trained by playing off two models against each other. 
The Generator learns to generate samples from the target distribution by taking a random vector as input (the GAN's ``inspiration") and transforming it into a valid image. 
The Discriminator is then tasked with distinguishing the fake, generated samples from real samples (i.e., natural images) from the training set. 
As the Discriminator gets better at separating real from fake images, the Generator is forced to generate increasingly more realistic samples, and vice versa.    
\end{svgraybox}

\subsection{Visualizations and model interpretability}

Knowing that a particular image is memorable does not necessarily tell you \textit{which} information is retained. 
Reasoning that not all image regions contribute equally to overall memorability, Khosla et al. \cite{khosla2012memorability} proposed a probabilistic model that assigns different weights to different regions based on six locally computed image features (gradient, saliency, color, texture, shape, semantics). 
This can be done automatically without the need for additional human annotations. 
Not only did accounting for local information in addition to global features improve the prediction of overall memorability scores, it also allowed for visualizing the weights as interpretable heat maps. 
These heat maps often emphasized regions depicting people as contributing most to the overall memorability, and plain backgrounds contributing the least.

\begin{figure}
\centering
\includegraphics[width=1\linewidth]{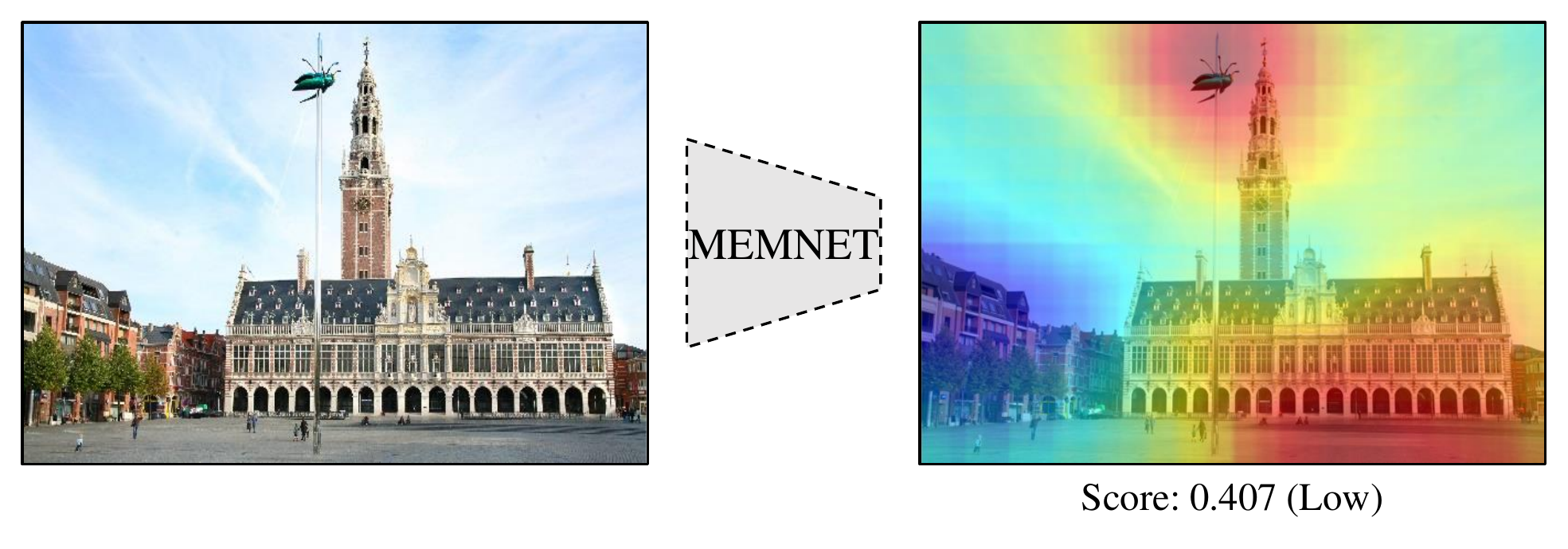}
\caption{MemNet prediction and memorability heat map for an example image. Left: Input image of Katholieke Universiteit Leuven \copyright Rob Stevens. Right: MemNet is a convolutional neural network fine-tuned to predict memorability scores \cite{khosla2015understanding}. The heat map is created by running the network over multiple, overlapping sub-regions and assigning each pixel a color based on the average predicted memorability of the regions that contain it. Here, it highlights the bell tower and bug sculpture as memorable regions.}
\label{fig:heatmap}
\end{figure}

As CNN models are particularly effective at predicting memorability, working out their internal representations yields valuable insights into which features result in high and low memorability scores. 
Khosla et al. \cite{khosla2015understanding} adopt different strategies to achieve this with MemNet. One is to sort the units within a layer of the network based on their correlation with the predicted memorability score, compute an average across the images that maximally activate the unit, and compare those averages. Another strategy, based on network dissection \cite{bau2017network}, is to identify image patches that highly activate individual units and then compare the activated patches from units that are positively versus negatively correlated with memorability.
% activation and compare what is in those patches between units that correlate positively versus negatively with memorability. This technique is based on network dissection \cite{bau2017network}. 
Finally, one can generate memorability heat maps by running the model on multiple, overlapping sub-regions of the image. Each pixel in the heat map represents the average memorability prediction across the sub-regions that contain it (Figure~\ref{fig:heatmap}). Together, these strategies revealed that MemNet tends to predict high memorability for images containing people or animals, busy images, and images with text. Open and natural scenes, landscapes and textured surfaces tend to result in lower memorability.

Another approach that facilitates model interpretation is visualizing what changes in an image when its model-predicted memorability increases or decreases, as in the GANalyze framework \cite{goetschalckx2019ganalyze}. For example, this framework showed that MemNet cares about the relative size of the main object in the image, with higher memorability predictions assigned to a more ``zoomed in'' image. While other techniques mostly revealed semantic features, GANalyze discovered important dimensions that are orthogonal to object class.

%Network dissection \cite{bau2017network, zhou2018interpreting, zhou2014object} can be used to work out the internal representations of CNNs. Specifically in the memorability space, Khosla et al. visualized memorability maps to show patches of images most positively and negatively correlated with the memorability of an image~\cite{khosla2015understanding}.

% ####################################################

\section{Memorability: from low-level to high-level features}\label{sec:features}

Since the original memorability experiments of Isola et al.~\cite{isola2011makes}, follow-up work has added further evidence to the observation that memorability is a robust property of images that is not amenable to simple explanation, but instead appears multifaceted (see~\cite{rust_understanding_image_memorability} for a review). It is correlated with scene category, but remains predictable when scene category is controlled for~\cite{bylinskii2015intrinsic,goetschalckx2019memcat}. It is correlated with the presence of certain elements - most notably, faces and people~\cite{baveye2016deep,bylinskii2015intrinsic,dubey2015makes,fajtl2018amnet,khosla2015understanding} - but remains predictable when those objects are not present~\cite{lu2018makes,9025769}. It cannot be explained by image aesthetics, popularity, or affective value either~\cite{cohendet2018annotating,goetschalckx2019ganalyze,isola2014what}. Moreover, human observers are bad at predicting what is memorable or forgettable~\cite{isola2014what}. So while no simple explanation for memorability has been proposed, in this section we will discuss some of the factors that contribute to image and video memorability, by applying the analysis tools and computational models from the previous section. In the concluding discussion (Section~\ref{sec:conclusion}), we hypothesize a unifying explanation to these results. 

\subsection{Low-level pixel features}

Dubey et al.~\cite{dubey2015makes} have found a weak positive correlation between memorability and the brightness and high contrast of objects. Using a GAN trained for memorability, Goetschalckx et al.~\cite{goetschalckx2019ganalyze} confirmed that brighter and more colorful images tend to be produced when optimizing for more memorable images. For instance, redder hues are produced when they are realistic (e.g., for ripe fruit). Lu et al.~\cite{lu2018makes} found that some HSV-based features could be used to predict the memorability of natural scene images without objects, though this predictive power is low ($\rho<0.30$).

For generic photographs, however, low level features (in the form of simple pixel statistics like color and contrast) are commonly either weakly correlated or uncorrelated with memorability~\cite{dubey2015makes,isola2011understanding,isola2014what,isola2011makes,9025769}. In general, perceptual features are not retained in long term visual memory~\cite{brady2011real,konkle2010conceptual}. %\TODO{Can we make more connections to Tim/Talia's past work?}

\subsection{Mid-level semantic features}

\textbf{Objects}. Multiple studies have commented on the increased intrinsic memorability of images and videos containing people, faces, and body parts~\cite{bylinskii2015intrinsic,dubey2015makes,khosla2015understanding,baveye2016deep,fajtl2018amnet,newman2020memento} and low memorability for landscapes~\cite{bylinskii2015intrinsic,goetschalckx2019memcat,isola2011makes,khosla2015understanding,newman2020memento}.
By using images with available semantic segmentations~\cite{li2014secrets}, i.e., images in which all or most of the objects have been delineated and annotated, Dubey et al.~\cite{dubey2015makes} showed that the memorability of an image is ``greatly affected by the memorability of its most memorable object" ($\rho=0.40$). For instance, animal, person, and vehicle were found to be the most memorable object classes, and images containing these objects were more likely to be memorable overall. These objects tend to dominate the focus and foreground of photographs, and are not commonly occluded~\cite{dubey2015makes}. In contrast, furniture was found to be the least memorable object category. The attention-based model of Fajtl et al.~\cite{fajtl2018amnet} was used to visualize regions of an image contributing most to memorability, and confirmed that these regions frequently contained people and human faces.

Objects are also important in video memorability prediction.
The importance of semantic features, including objects, actors, and the actions that involve them, has increasingly led to their inclusion in video memorability models \cite{newman2020memento,cohendet2019videomem,cohendet2018annotating}.
The semantic embedding model from VideoMem~\cite{cohendet2019videomem} was based on a video captioning model, and SemanticMemNet~\cite{newman2020memento} includes a branch to explicitly generate video captions to encourage features that encode semantic information.

\textbf{Object interactions}. Object interactions are key to the memorability of individual objects as well as the memorability of the image containing them. In particular, objects that are out of context with respect to the other items in a scene make for particularly memorable images~\cite{bylinskii2015intrinsic,standing1973learning}. Also, as the pure number of objects in an image increases, competition for attention decreases the memorability of even the most memorable object classes, like animals and vehicles~\cite{dubey2015makes}. Interestingly, the memorability of people in images is least sensitive, compared to other object classes, to the presence of other objects in an image. 

\textbf{Saliency and eye fixations}. Multiple studies have explored the connection between image memorability and saliency~\cite{dubey2015makes,khosla2015understanding,mancas2013memorability,9025769,shekhar2017show,lu2018makes, celikkale2015predicting, akagunduz2019defining}. Mancas and Le Meur~\cite{mancas2013memorability} ran computational saliency models on images and found that the most memorable images have localized regions of high saliency, while the least memorable images do not. Khosla et al.~\cite{khosla2015understanding} reported that more memorable images tend to have more consistent human eye fixations. Dubey et al.~\cite{dubey2015makes} similarly found a large positive correlation ($\rho = 0.71$) between fixation count on an object in an image and that object's memorability. Part of this trend is driven by the fact that objects that are not fixated at all are not remembered. The other aspect is that images that contain more close-ups or larger objects will tend to have more consistent fixations clustered on those objects (also see \emph{image composition} below). As the number of objects in an image increases, it becomes significantly harder to predict the memorability of objects using fixation counts alone. While computational saliency is intended to simulate human attention and serve as an approximation of eye fixation patterns, they are not the same thing. Computational saliency can be used as a replacement to other low-level features in predicting memorability~\cite{mancas2013memorability}. Lu et al.~\cite{9025769} found that the performance of a saliency model on saliency tasks is not correlated with its performance on memorability prediction, affirming that computational saliency in this regard should be viewed as a pre-computed combination of other low-level features rather than an independent high-level semantic feature. Alternatively, rather than being used directly as an input feature, saliency maps can also be used to spatially pool other low-level input feature and improve memorability predictions \cite{celikkale2015predicting}.

\textbf{Image composition}. More memorable images tend to focus on a key object or image region, and to center it in the photograph while maintaining a homogeneous background. For instance, a photograph of a puppy is more memorable when it is a close-up and the puppy occupies a larger portion of the photograph~\cite{goetschalckx2019ganalyze}. This is similar to Dubey et al.'s~\cite{dubey2015makes} finding that if fewer objects compete for attention, then a single object becomes quite memorable, and the image as a whole becomes memorable by extension. Kim et al. \cite{kim2013relative} showed that features computed based on the relative sizes of objects, their centeredness, and the unusualness of their size given the overall semantic class are strong predictors of memorability. When GANalyze \cite{goetschalckx2019ganalyze} modifies a seed image to become more memorable, it typically makes the main object larger and more centered, simplifies the image, and reduces clutter. The authors furthermore note that ``more memorable images have more interpretable semantics''. In an additional experiment, they show that when GANalyze is trained to modify object size directly, the image variants with the larger object size are indeed more likely to be remembered by participants in a memory task. However, the effect is not as strong as when memorability was targeted as a whole, indicating that memorability is more than object size alone.   %Goetschalckx et al.~\cite{goetschalckx2019ganalyze} found that simpler, less cluttered, and better structured images are more memorable and that ``more memorable images have more interpretable semantics".
Simpler, more orderly shapes are more memorable than a disarray or lack of structure. Fajtl et al.'s~\cite{fajtl2018amnet} visualizations confirm that images with higher memorability display more concentrated peaks of memorable content, whereas lower memorability images tend to have memorability more distributed across the image. Finally, Mancas and Le Meur \cite{mancas2013memorability} proposed a feature to capture the presence of strong contrasted structures in an image, as opposed to small details and cluttered backgrounds, and found that it positively predicted memorability.

\begin{figure}
\centering
\includegraphics[width=1\linewidth]{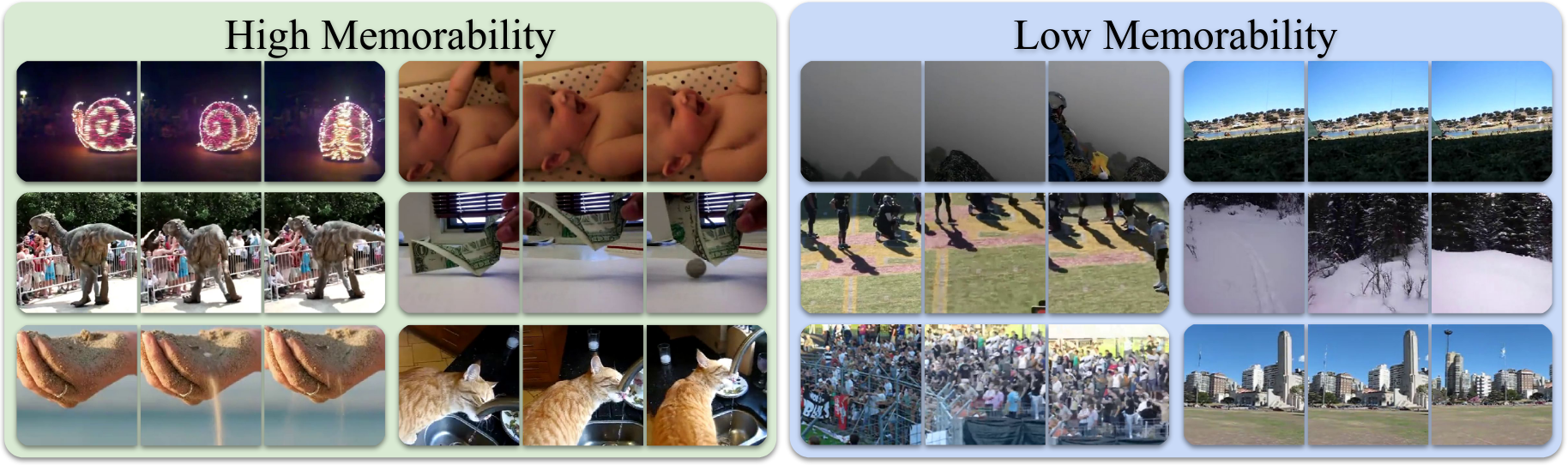}
\caption{High and low memorability videos from the Memento dataset \cite{newman2020memento}. Like images, memorable videos tend to contain people, faces, and body parts, while forgettable videos are more likely to be of natural and otherwise static scenes.}
\label{fig:memento}
\end{figure}

\textbf{Motion and action features}.
Using videos as stimuli has facilitated the analysis of the effects of motion on memory.
Newman et al.~\cite{newman2020memento} evaluated the effectiveness of different feature extractors that did or did not explicitly encode motion for predicting video memorability. 
They found that a simple image-based extractor that operated only on static frames performed as well as a video-based extractor, but that combining information from static frames, video, and optical flow produced the best results. 
They also observed that low-memorability videos are often static compared to high-memorability videos (Figure~\ref{fig:memento}).

\subsection{High-level and contextual features}

%\TODO{Consider changing the structure a little bit? Adding some suggestions in comments (Lore)}

% I'm not sure if I'd place things like aesthetics and emotional valence under "contextual" features. And I'm not sure if I would bundle everything under "extrensic". Shouldn't we also include a paragraph on regular semantic stuff, category differences etc.? Those are very strong predictors. Maybe we can do:

%4.3 High-level features
%   - Scene and object semantics, cite things like \cite{khosla2015understanding, isola2011understanding,9025769, isola2011makes} and more
% - Abstract, subjective properties
%       * Aesthetics \cite{isola2011makes, khosla2015understanding, goetschalckx2019ganalyze}
%       * Emotions \cite{khosla2015understanding, goetschalckx2019ganalyze, baveye2016deep, celikkale2015predicting}
%       * Popularity 
%       * Interestingness \cite{isola2011makes,gygli2013interestingness}

%4.4 Contextual features
% Discuss the major role of distinctiveness here, definitely \cite{bylinskii2015intrinsic}. Probably \cite{lukavsky2017visual}, maybe \cite{goetschalckx2019get} (behavioral distinctiveness measure, so not automatically computed, but Fig. 11 might be relevant). And some other atypicality, unusualness work \cite{kim2013relative,9025769} (there might be more?)

% I think maybe some of the things I wrote in the discussion of my PhD dissertation might be relevant. It doesn't need to be cited, but just to share some ideas (e.g., on the difference between primary and secondary distinctiveness). See section 7.2.1 and 7.2.3: https://lirias.kuleuven.be/retrieve/552172

\textbf{Aesthetics}. Aesthetics 
% has been found to be 
is distinct from memorability, with little to no correlation in the LaMem dataset \cite{khosla2015understanding,jing2016predicting} and a weak, but negative correlation in the SUN dataset~\cite{isola2014what}. Participants tend to rate images depicting nature as most aesthetic (e.g.,~coast and lake scenes), yet natural landscapes typically score low on memorability. Modifying an image to increase its aesthetics (while keeping semantic class constant) does have a positive effect on memorability, albeit a rather small one ~\cite{goetschalckx2019ganalyze}. Goetschalckx et al.~\cite{goetschalckx2019ganalyze} furthermore confirmed that optimizing images for memorability and optimizing images for aesthetics leads to different image manipulations.

\textbf{Emotions}. %It is known that e
Emotionally salient objects are memorable \cite{bradley1992remembering,buchanan2002role,cahill1995novel}. Hence, different studies have considered the role of emotional features in predicting image memorability. In the LaMem dataset, images evoking negative emotions (e.g., disgust, anger, fear) were overall more memorable than those evoking positive emotions (e.g., awe, contentment), with amusement being a memorable exception \cite{khosla2015understanding}. Isola et al. \cite{isola2011understanding} found that images described as peaceful are typically not well remembered. Having a subject in a photo with more pronounced and expressive eyes, on the other hand, seems to increase the memorability of a portrait \cite{goetschalckx2019ganalyze}. In \cite{jing2016predicting}, a sentiment attribute inspired by eight basic emotions was the best memorability predictor among other high-level attributes (e.g., aesthetics). Other work that included emotional features combined with other high-level features (e.g., object categories) also demonstrated a predictive ability for these features \cite{isola2011understanding, celikkale2015predicting}. Finally, Baveye et al.~\cite{baveye2016deep} recommend that memorability datasets should have the appropriate emotional feature distribution (i.e., be balanced), based on the observation that emotionally negative images have more predictable memorability than neutral or positive ones. 
% Lore: removed valence from the title as I thought it might be too specific. Refers to where on the negative - positive spectrum an emotion is situated, but there is also the arousal spectrum.

\textbf{Popularity and interestingness}. Intuitively, one might expect memorable images to be 
% those that are 
popular or interesting. However, studies show that memorability does not reduce to either of those attributes. The top 25\% most memorable LaMem images indeed are more popular, as determined by their log-normalized view count on Flickr. This is likely because they stand out in some way. However, there is little difference between all other images \cite{khosla2015understanding}. When asking participants to judge the interestingness of an image in the SUN dataset, those judgments correlate negatively with memorability, but only weakly \cite{isola2014what, gygli2013interestingness}.

\textbf{Scene category}. Perera et al.~\cite{perera2019image} empirically showed that a network originally trained on scene classification transfers more appropriate features for memorability prediction than a network trained on object classification, although a network trained on both performs best~\cite{zhou2014learning,khosla2015understanding,zhou2017places}. This seems to indicate that scene category drives a lot of the predictive power in image memorability. Indeed, Isola et al.'s original paper \cite{isola2011makes} reports that scene category alone is highly predictive of memorability ($\rho=0.37$). Lu et al.~\cite{9025769} obtain a similar result ($\rho=0.38$) across their dataset of natural outdoor scene images, LNSIM. Their images do not contain any salient objects (people, animals, man-made structures, etc.) and yet they still demonstrate high inter-observer consistency ($\rho=0.78$). Finally, 43\% of the variance in the MemCat dataset was captured by which of five broad categories an image belonged to \cite{goetschalckx2019memcat}.
However, scene category is far from the whole story. Memorability scores in the MemCat dataset were still consistent across observers within each of the five categories separately \cite{goetschalckx2019memcat}. This still holds in cases where the scene category is even more strictly controlled for, and in this case, objects and their distributions help to drive memorability~\cite{bylinskii2015intrinsic}. Finally, GANalyze \cite{goetschalckx2019ganalyze} is able to increase image memorability while keeping the scene category constant.
% In other words, the scene category of an image drives

Despite all the work showing that memorability has a strong intrinsic component, features extracted automatically from image pixels account for only about half the total variance observed in the memorability scores. The rest may be due to individual or context factors, which we can refer to as the extrinsic factors~\cite{bylinskii2015intrinsic}. 

\textbf{Observer attention}. Unlike saliency, which approximates the attention patterns of a population in aggregate, Bylinskii et al.~\cite{bylinskii2015intrinsic} considered whether an individual's pattern of eye movements
% knowing where a particular individual looked could be 
could predict whether that individual would remember an image. The model was an SVM trained to separate eye fixation patterns on a target image from eye fixation patterns on other images. This model achieved approximately 60\% accuracy at predicting if a given image would be remembered by an individual based on their eye fixation patterns alone. Using this person-specific information was more accurate than the population-level predictor (average memorability scores), especially for the images in the mid-memorability range.

\textbf{Contextual features}. Images that stand out from their image context (i.e., other images presented in the memory task) or differ from our expectations based on an internal model of the world, are more likely to be remembered.  Bylinskii et al.~\cite{bylinskii2015intrinsic} were able to show this by operationalizing contextual distinctiveness in terms of how unlikely an image's features are in the feature distribution defined by the image context. Similarly, Lukavsk{\'{y}} and D{\v{e}}cht{\v{e}}renko \cite{lukavsky2017visual} found that people remember images better if they are far away from their nearest neighbors in a conceptual representational space. Goetschalckx et al. \cite{goetschalckx2019get} compared these two automatic distinctiveness measures with a perceived distinctiveness measure based on participants' ratings, as well as a measure of an image's atypicality for its abstract scene category. All four variables were significantly intercorrelated. Perceived distinctiveness predicted memorability scores best.

%Related results include that 
Furthermore, of the 60k images in the LaMem dataset, those originally sampled from the Abnormal Objects dataset~\cite{saleh2013object} are extremely memorable \cite{khosla2012image}. Landscape images in Lu et al.'s dataset \cite{9025769} belonging to unusual categories (as indicated by the category name having a low word frequency in language) tended to be more memorable. Finally, Kim et al. \cite{kim2013relative} report that accounting for how unusual object sizes are given their class can improve memorability predictions.

Lastly, another way in which context matters is through effects of image sequences and presentation order, as demonstrated by Perera et al.~\cite{perera2019image} This confirmed earlier results from \cite{bylinskii2015intrinsic}.

%\TODO{TODO: discuss \cite{lukavsky2017visual,goetschalckx2019get,kim2013relative,9025769} --> Done! (I think - Lore)} %showed that images that are atypical category exemplars are predictably most memorable.
% Lore: I'm not so convinced that Kim et al. unusual size feature contributed much beyond their other feature though ... 

% ####################################################

\section{Applications: from summarization to creation}\label{sec:applications}

As computational models have recently approached human-level performance at predicting what people will find memorable, the doors have opened to applications that could benefit from an estimate of human memorability.
% As humans are quite consistent in what they find memorable, and computational models have recently approached human-level performance at predicting memorability, the doors have opened to applications that can benefit from these results. 
In this section, we provide a taste of some of these applications. Making more of them a reality depends on additional progress to be made along the future research directions proposed in the following section.

\textbf{Filtering visual streams}. With a growing stream of information and an increase in low-cost, low-power image/video capture devices (phones, GoPros, internet of things, etc.), filtering through content to find the nuggets worth saving becomes increasingly tedious and time-consuming. Such a task could be outsourced to computational agents armed with an appropriate filtering criteria. Here, memorability can play a significant role, as a high-level image property that bundles together low-level, semantic, and contextual features. Which of the dozens of nearly-identical photos should be saved for later? Which frames or snapshots can represent the contents of a video in summary form? Current computational models of memorability could be used for ranking and effectively filtering visual content.

\textbf{Assistive goggles}. Imagine an automated tutor that reminds the wearer of the identities of the most forgettable objects and people, and either coaches the user's memory or presents the labels in an augmented reality layer. Augmented Reality (AR) is a technology that integrates digital graphics and virtual objects into a display of the real world. It is already being used for training and education scenarios in medicine~\cite{barsom2016systematic}, construction~\cite{li2018critical}, driving~\cite{gabbard2014behind}, and K-12 education~\cite{holstein2018cognitive_augmentation_teachers}. As a potential tool for augmenting the human memory of the average person, it holds big promises for the future. 

\textbf{Photography aide}. Consumer cameras are increasingly being upgraded with machine-learning based functionalities for guiding the user towards capturing better-quality shots. These include new tools to help users automatically select a better portrait angle or orientation for a photograph~\cite{e2020adaptive,ma2019smarteye}. A more sophisticated aide could guide the photographer to select a more memorable shot during real-time capture, or to re-position, adjust, and remove elements either before or after a picture is taken, by optimizing for memorability. 

\textbf{Effective communication}. Key players in the education space (e.g., educational content producers, intelligent tutoring systems, etc.) could benefit students by presenting content in a more memorable way. For instance, prior work has looked at data visualizations and their ability to effectively communicate information via memorable presentations of data and title wording~\cite{newman2018effects,borkin2015beyond,xiong2019biased,kong2018frames,kong2019trust}. Armed with similar insights, a marketing or advertising effort could make use of predictions of how to arrange the objects in a scene to make the target product or message stand out in a memorable way. Along the same lines, a tourist agency could design more memorable experiences and increase re-visitations~\cite{HungTouristm,kleinlein2019predicting}. 

\begin{figure}
\centering
\includegraphics[width=1\linewidth]{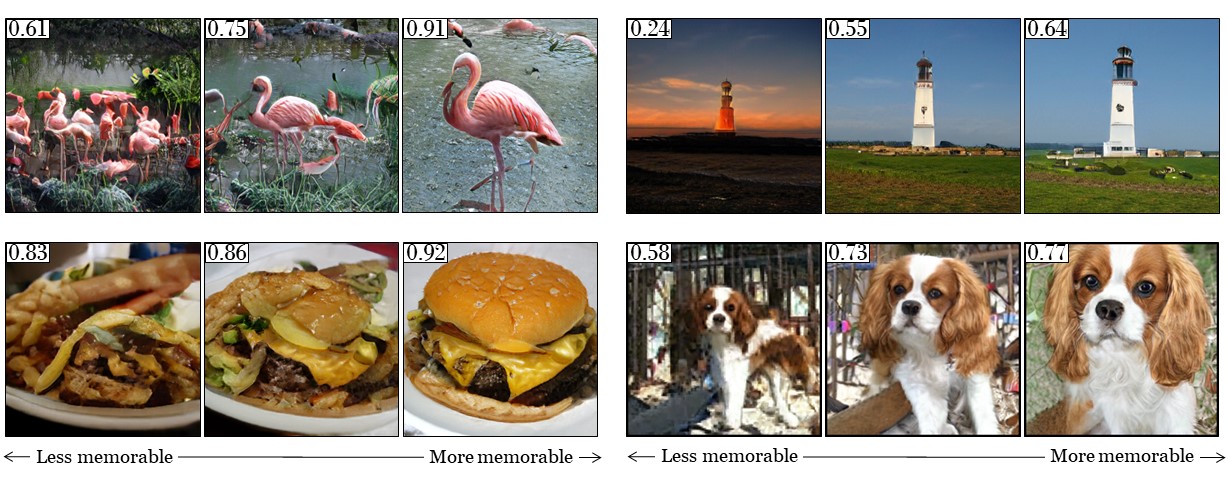}
\caption{More or less memorable images generated by the GANalyze framework \cite{goetschalckx2019ganalyze} with computed memorability scores as insets. More memorable images tend to focus on a large, central key object while reducing background clutter.}
\label{fig:ganalyze}
\end{figure}

\textbf{Manipulating memorability}. What if an app could make your holiday pictures extra memorable or the picture on your resume more likely to be remembered by a recruiter? Our growing ability to predict and understand memorability has led researchers to think about the possibility of automatically manipulating an image’s memorability. For example, Khosla et al. \cite{khosla2012image} speculated that their probabilistic model of the memorability of images and their sub-regions \cite{khosla2012memorability} could offer a starting point for memorability manipulation work. Deep style transfer has also been put forward as a way to boost an image’s memorability score \cite{siarohin2017make,siarohin2019increasing}. By transferring the style of a seed image onto the original image, the modified version also tends to have a more abstract, artistic flavor to it. Furthermore, the memorability of a face image can be manipulated successfully using warping techniques, all while maintaining the identity of the face \cite{khosla2013modifying}.
More recent work has turned to GANs for this purpose \cite{goetschalckx2019ganalyze, Sidorov_2019_CVPR_Workshops}. While Goetschalckx et al.’s~\cite{goetschalckx2019ganalyze} GAN framework successfully increased and decreased the memorability of images (Figure~\ref{fig:ganalyze}), this was only possible by using seed images that were GAN-generated to begin with - i.e., within the latent space of the GAN. However, recent success in GAN inversion (i.e., projecting a real image into a GAN’s latent space) \cite{zhu2020indomain, abdal2019image2stylegan, bau2019ganpaint, anirudh2020robust, creswell2019inverting} suggests that it might be possible to extend these results to real, user-supplied imagery.
While automatically boosting an image’s memorability is an exciting possible direction, similar to automatically beautifying portraits, one must consider the potential concerns about image authenticity that it raises.

%While Goetschalckx et al.~\cite{goetschalckx2019ganalyze} used a GAN framework to increase and decrease the memorability of images, this was only possible by using seed images that were GAN-generated to begin with - i.e., within the latent space of a GAN. So far, no work has yet modified a user-supplied, real image, to make it more/less memorable. While this is an exciting possible direction, similar to automatically beautifying portraits, it does raise issues about image authenticity.

%Editing face memorability using warping techniques~\cite{khosla2013modifying} and GANs~\cite{Sidorov_2019_CVPR_Workshops}; memorability editing using style transfer~\cite{siarohin2017make}; inception~\cite{khosla2012image}.

% ####################################################

\section{Future directions}\label{sec:future}

% Similar to the success of other image-level prediction tasks~\cite{sharif2014cnn,zhou2014learning}, appropriately-trained 
Convolutional neural networks can now closely predict the average memorability score of an image. 
Perera et al.~\cite{perera2019image} present a model that is able to reach human-level performance on the LaMem dataset~\cite{khosla2015understanding}, the largest memorability dataset to date.
They pose the question: is memorability prediction solved? 

Despite the success of computational models of image memorability, there are still aspects of memorability prediction and utilization that merit further work. 
We discuss some of these directions below.

\begin{figure}
\centering
\includegraphics[width=0.8\linewidth]{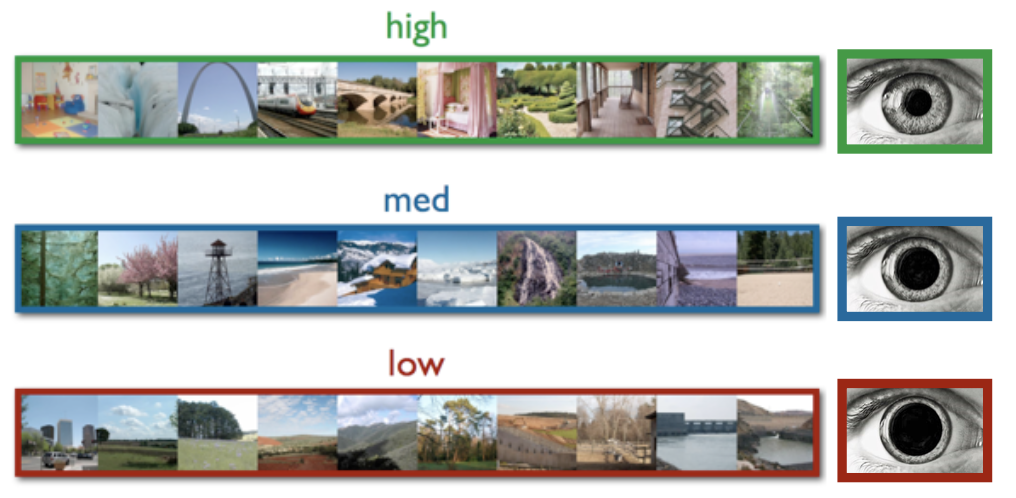}
\caption{Pupillometry analysis shows that at recall, images with the lowest memorability cause a dilation in pupil size relative to images with the highest memorability, on an individual basis~\cite{Vo2017}.}
\label{fig:pupilometry}
\end{figure}

\textbf{Customized predictions}.
Producing the average memorability score of an image overlooks possible variability in the population~\cite{bylinskii2015intrinsic,perera2019image}.
Variability may exist among subpopulations or specialists, people from different cultures and environments (i.e., LaMem uses US crowdworkers on Amazon's Mechanical Turk platform), and among individuals.
Drilling down to the individual level for customized applications of memorability is an exciting prospect. Can we predict whether a particular image will be memorable to a particular individual? While Bylinskii et al.~\cite{bylinskii2015intrinsic} and V{\~o} et al.~\cite{Vo2017} showed that eye movements and pupillometry (Figure~\ref{fig:pupilometry}), respectively, could be used for individualized memorability predictions, predictions are not yet accurate nor robust enough to be deployed in applications and devices. Deep learning based face and eye tracking technology~\cite{krafka2016eye,Zhang_2015_CVPR,park2020towards} and other automatically-collected physiological measurements~\cite{papoutsaki2016webgazer,qian2018remotion} may provide a way forward.

\textbf{Retention over different time intervals}. %Past work has explored to what extent image memorability is stable over shorter or longer intervals than the few minutes typically used in the memorability games. For instance, Isola et al.~\cite{isola2013makes} and V{\~o} et al.~\cite{Vo2017} showed that even after a few seconds, some images are already consistently more  forgotten, and that although memorability ranks are conserved, some images decline in memorability at a higher rate than others. \TODO{TODO: Fill in specifics for the above, and talk about retention over longer time intervals, and specifically, the future directions of what has yet to be explored.}
The memorability games used to create most of the available datasets (Table~\ref{tab:datasets}) typically measure memorability over intervals of a couple of minutes. By varying the number of images presented between a target image and its repeat (i.e., the ``lag''), past work has explored to what extent image memorability is stable over shorter and longer intervals \cite{isola2011makes, Vo2017}. Goetschalckx et al. \cite{goetschalckx2018longer} used a more traditional long-term memory task with a separate study and test phase that allowed them to study intervals as long as a week. In addition, different mathematical formulations have been proposed to compute memorability scores based on responses collected at varying lags \cite{khosla2015understanding, newman2020memento}.

An image's memorability score naturally decays over time, but even after a few seconds, some images are already consistently more likely to be forgotten \cite{isola2011makes, Vo2017}. The decay is best described by a log-linear function \cite{isola2011makes, goetschalckx2018longer} and is slower for more memorable images \cite{Vo2017, goetschalckx2018longer}. Despite changes in raw scores over time, memorability ranks are largely conserved \cite{isola2011makes, goetschalckx2018longer}, though more substantial differences might appear when comparing intervals significantly differing in length \cite{goetschalckx2018longer}.

Video memorability studies have likewise compared different intervals. 
Cohendet et al. \cite{cohendet2019videomem} report a significant, albeit moderate, correlation between video memorability rankings measured after a couple of minutes and after 24 to 72 hours. The correlation was likely attenuated by the low number of responses collected per video, however. 
Cohendet et al. \cite{cohendet2018annotating} examined even longer durations by asking study participants about well-known films that they had seen weeks to years previously. 
They find an inter-observer consistency score of 0.57, which is lower than the values of most memorability studies but still significant (see Table \ref{tab:datasets}). 
The lower consistency could be due to the long delay between initial and repeat viewing, the relatively low number of participant annotations per video, or the atypical experiment design where participants self-reported the time of their first viewing.
Similar to images, the Memento10k videos \cite{newman2020memento} also showed slower decay if they had higher memorability scores to begin with. Unlike with images, the decay was best described by a linear function. Because the authors examined relatively short delays (up to ten minutes between viewings), it is possible that at longer delays, a log-linear trend may be more appropriate.
The authors proposed a computational model to simultaneously capture a video's memorability and decay rate.
% Compared to the number of computational models that have been proposed to predict memorability scores for images, those available for videos are much fewer. 
Models that attempt to predict memorability as a function of time interval (how memorable will this image be in a day? a week? a year?) or by directly predicting the decay rate for an image or video \cite{newman2020memento} are rare. These questions deserve more attention.

%Building on the work by Newman et al. \cite{newman2020memento}, future work could look to model image memorability while including time interval as a predictor and studying its interactions with predictive image features. For example, which images do climb or drop in the ranking with time and what sets them apart? 
% Lore: Was this along the lines of what you are looking for with future directions? I'm not sure if I did good with this suggestion, because it kind of emphasizes differences whereas the general finding is stability over time.

%Lore: These are the studies I could think of from the top of my head. There is another one comparing memorability of scrambled images at short and long intervals, but it's not published (saw it at VSS) and I vaguely remember there being a problem with the long term taks being too difficult. I didn't spend too many words on the Cohendet study because I feel that the the low number of responses is problematic (like they say something about how it's more difficult for a model to predict long-term scores, but at the same time their long-term scores were based an way fewer responses)

%\TODO{Break this up into sections and talk about different possible directions: specialized populations, retention over longer time periods, other stimuli times, text-based memorability, etc.}

\textbf{Alternative media}. 
Much of the work in computational memorability thus far has focused on images. Compared to the number of computational models that have been proposed to predict memorability scores for images, those available for videos are much fewer \cite{cohendet2018annotating, cohendet2019videomem, newman2020memento} and they do not yet reach human-level performance (see Table \ref{tab:vidmodels}).
As such, perfecting the nuances of video memorability prediction is an important future direction.

Furthermore, the prior work discussed in this chapter has investigated purely visual media (images and videos without sound). Future work can explore the influence of auditory features (voices, sound effects, music, etc.) on memorability, independently of or in combination with visual features. Some work has additionally been done in the space of word memorability~\cite{mahowald2018words,madan2020exploring}, with findings that reinforce those in visual domains, in that distinct, easily-visualizable, non-ambiguous words are the most memorable~\cite{mahowald2018words}. Mahowald~\cite{mahowald2018words} found that animacy was most correlated with word memorability, and further that words that were rated as related to `danger' and `usefulness' were more memorable. The memorability of combinations of words, in the form of descriptions or titles, has yet to be studied, especially in the context of conveying information, and making messages or data ``stickier"~\cite{newman2018effects,borkin2015beyond,xiong2019biased,kong2018frames,kong2019trust}. 

\section{Concluding discussion}\label{sec:conclusion}

Putting it all together, what is the ``magic sauce'' of memorability? What is the common thread that ties the findings from past memorability studies together, across photographs, visualizations, videos, and words? 

We propose that content is memorable if it has a high \textit{utility of information}. We remember that which surprises us, that which contradicts our current model of the world and of events, or that which is likely to be relevant or useful in the future. Remembering a surprising (or dangerous, untrustworthy, etc.) event or person will prepare us better for future encounters, and for adjusting our world view accordingly. This is why stimuli that are distinct, relative to their contexts, are memorable. 

% The context for a stimulus could be interpreted to include only the other stimuli that appear within a given viewing window (e.g., the neighboring stimuli, the stimuli that make up the experimental sequence, etc.) or all other stimuli from the same category (e.g., same scene type, similar semantics, etc.). There is content that is memorable only within certain collections or datasets, and other content that is memorable nearly independently of its context. 
The benefit of large, diverse memorability datasets with many observers is that they capture the robustly memorable and forgettable content.
These studies tell us something about the human condition: What has the highest utility of information to any observer? Universal trends include emotional/affective stimuli, unexpected actions, social aspects, animate objects (human faces, gestures, interactions, etc.), and tangible (small, manipulable) objects. Memorability is not about aesthetics or low-level visual features like color or contrast. Rather, memorability captures the higher-level properties of semantics (objects and actions) and composition (layout and clutter) in an image or video. 

As a demonstrated image-computable attribute, memorability has the capability of being used as a powerful image descriptor or feature representation for downstream tasks that depend on image understanding and selection. Cognitively-inspired computation is still in its relative infancy. Memorability paints a promising path into the A.I. future. 

%\subsection{Open questions}

%\textbf{Specialized populations.} Airplane cockpits are among the most forgettable of scene categories, as all the dashboards blend together for the average observer, lacking distinctiveness~\cite{bylinskii2015intrinsic}. However, the placement of those controls has meaning for those who work with them daily, so memorability for individual cockpit scenes is expected to be higher among pilots. Similar observations should hold for different specialist populations. Image memorability, in the format presented in this chapter, has not been extensively studied with different specialized populations of participants. 

\section{Acknowledgements}
We thank the Vannevar Bush Faculty Fellowship Program of the ONR (N00014-16-1-3116 to A.O.). Thank you also to Wilma Bainbridge for her insightful comments on prior versions of this chapter, as well as to the other editors of this collection.

\bibliographystyle{spmpsci}
\bibliography{references}

\end{document}